\documentclass[conference]{IEEEtran}
\IEEEoverridecommandlockouts
\usepackage{cite}
\usepackage{amsmath,amssymb,amsfonts}
\usepackage{algorithm}
\usepackage{algpseudocode}
\usepackage{leftidx}
\usepackage{setspace}
\usepackage{graphicx}
\usepackage{textcomp}
\usepackage{mathtools}
\usepackage{xcolor}

\newtheorem{definition}{Definition}
\newtheorem{theorem}{Theorem}
\renewcommand{\S}[1]{\ensuremath{\mathbf{#1}}} 
\newcommand{\Ip}{\leftidx{^I}{p}}
\newcommand{\op}{\leftidx{^{opt}}{p}}
\def\BibTeX{{\rm B\kern-.05em{\sc i\kern-.025em b}\kern-.08em
    T\kern-.1667em\lower.7ex\hbox{E}\kern-.125emX}}
\begin{document}

\title{Similarity-based transfer learning of decision policies
\thanks{Partial support of the project M\v{S}MT LTC18075 is acknowledged.}
}

\author{\IEEEauthorblockN{Eli\v{s}ka Zugarov\'{a}}
\IEEEauthorblockA{\textit{Adaptive System Department} \\
\textit{Institute of Information Theory and Automation} \\
\textit{The Czech Academy of Sciences}\\
Prague, Czech Republic \\
eliska.zugarova@gmail.com}
\and
\IEEEauthorblockN{Tatiana V. Guy,\IEEEmembership{ Senior Member, IEEE}}
\IEEEauthorblockA{\textit{Adaptive System Department} \\
\textit{Institute of Information Theory and Automation}\\
\textit{The Czech Academy of Sciences}\\
Prague, Czech Republic \\
guy@utia.cas.cz}
}

\maketitle

\begin{abstract}
We consider a problem of learning decision policy from past experience available. 
Using the Fully Probabilistic Design (FPD) formalism, we propose
a new general approach for finding a stochastic policy from the past data.
The proposed approach assigns the degree of similarity to all of the past closed-loop behaviors. 
The degree of similarity express how close the current decision making task is to a past task. Then it uses Bayesian estimation to learn an approximate optimal policy, which comprises the best past experience.
The approach learns decision policy directly from the data without interacting with any supervisor/expert or using any reinforcement signal.
The past experience may consider a decision objective different than the current one. Moreover the
past decision policy need not to be optimal with respect to the past objective.
We demonstrate our approach on simulated examples and show that the learned policy achieves better performance than optimal FPD policy whenever a mismodeling is present.
\end{abstract}


\section{Introduction}
Learning from the past experience that mimics the expert's/teacher's decisions or behaviour (for instance imitation learning \cite{Wu_etal:19}, apprenticeship learning, \cite{AbbNg:04}) become popular in the recent years.

Successful applications, like natural language processing \cite{Cha_etal:15}, autonomous micro aerial vehicles \cite{Kau_etal19}, mimicking human body language in games \cite{Lev_etal:09}, support the interest in developing these approaches.
Often expert's behavior are very complex (for instance car driving) and
can hardly be represented via a set of feasible algorithms. On other hand
demonstrating a desired behavior may be easy for the expert but designing algorithms
imitating this behaviour is mostly difficult. Even when designed, these algorithms have a limited ability to generalise past experience and to find an optimal decision policy under new circumstances (for instance change of the system or new decision making preferences).
Another problem that to reach a high quality decision policy the existing algorithms may require a large amount of demonstration/expert data in long-horizon problems \cite{RosBag:2010}. 

Many successful approaches rely on querying the expert, thus becoming dependent on an expert's feedback which is often not feasible or restricted by application domain.
Generally learning targeted, sequential decision-making behavior is quite difficult for a general problem, when the resulting algorithm must often reason about the long-term consequences of the currently chosen actions.

In the paper we propose a new general approach for finding a stochastic policy from the past data (either generated by expert or not).

The proposed approach uses the Fully Probabilistic Design (FPD) formalism \cite{Kar:06Ful},\cite{Kar:20}. We compute value of similarity that reflects how much each past closed-loop behavior matches to the targeted closed-loop behavior. Then, using these similarity values the optimal decision policy is learned from all of the past data.
The resulting decision policy thus comprises the best experience obtained in the past. The approach learns decision policy directly from the data without interacting with any supervisor/expert or using any reinforcement signal. The past experience may consider a decision objective different than the current one. Moreover the past decision policy need not to be optimal with respect to the past objective.
We demonstrate our approach on simulated examples and show that the learned policy achieves better performance than optimal FPD policy whenever a mismodeling is present.

The paper outline is a follows. The next section introduces notations and notions, recalls necessary information about the Markov Decision Processes and Fully Probabilistic Design.
Section~\ref{sec:TL} formulates and solves similarity-based transfer learning. Section~\ref{sec:Illustrative_experiments} describes the algorithm and experiments performed; outlines and discusses the obtained results. Section~\ref{sec:Conclusions} provides concluding remarks.

\section{Preliminaries} 
\subsection{General Notation}
\begin{itemize}
  \item $\mathbb{N}$ and $\mathbb{R}$ stand for sets of natural and real numbers, respectively.
  \item Sets of values of discrete random variables are denoted by bold capital letters, i.e. $\S{X}$ is a set of values $x$.
  \item $|\S{X}|$ represents the cardinality of set $\S{X}$.
  \item Value of variable $x$ at discrete time $t\in\mathbb{N}$ is denoted as $x_t$.   
  \item  $p(x)$ denotes probability mass function of a discrete random variable $x$ onwards referred to as probability function;
  \item $p(x|y)$ is the conditional probability of a discrete random variable $x$ conditioned on random variable $y$;
  \item $\mathrm{E}[x]$ is expectation of a random variable $x$ and $\mathrm{E}[x|y]$ denotes conditional expectation of a random variable $x$ conditioned on a random variable $y$.
\end{itemize}

\subsection{Markov Decision Process}

Markov Decision Process (MDP) \cite{Put:94} is a framework widely used for sequential decision-making problems. It serves to model an agent that interacts with a system by deliberately choosing actions to achieve its objectives expressed in the form of a reward function.
\begin{definition}[MDP] A finite-horizon discrete-time fully observable Markov Decision Process is a tuple $\lbrace\S{T},\S{S},\S{A},p,r\rbrace$, where
\begin{itemize}
  \item []$\S{T}=\lbrace 1, 2, ..., N\rbrace, N\in \mathbb{N}$, is a set of decision epochs,
  \item []$\S{S}$ is a discrete finite set of all achievable system states,
  \item []$\S{A}$ is a discrete finite set of all possible actions of the agent,
  \item []$p:\S{S}\times\S{A}\times\S{S}\rightarrow [0,1]$ is a transition probability function that models the evolution of the system, $p(s_t|a_t,s_{t-1})$ is  the probability that the system moves from state $s_{t-1}\in\S{S}$ to state $s_t\in\S{S}$ after action $a_t\in\S{A}$ is taken, $t\in\S{T}$,
  \item []$r:\S{S}\times\S{A}\times\S{S}\rightarrow \mathbb{R}$ is a reward function; $r(s_t,a_t,s_{t-1})$ is the immediate reward the agent receives after taking action $a_t\in\S{A}$ in state $s_{t-1}\in\S{S}$ and prompting the system to move to state $s_t\in\S{S}$, $t\in\S{T}$.
\end{itemize}
\end{definition}
The system transition is ruled by the Markov property \cite{Put:94}, which means that it depends only on the last system state and the chosen action.

The agent wants its action selection to lead to maximum possible reward. The optimal behavior is determined by choosing an action maximizing the total expected reward at each decision epoch. The action selection is ruled by a decision policy, which is a sequence of decision rules and can be expressed as
\begin{displaymath}
  \big\{p_t(a_t|s_{t-1})|a_t\in\S{A},s_{t-1}\in\S{S}\big\}_{t=1}^{N}.
\end{displaymath}
Each decision rule is a conditional probability function over the action set.

\subsection{Fully Probabilistic Design}\label{sec:FPD}
The Fully Probabilistic Design (FPD) framework \cite{Kar:96,Kar:06Opt,Kar:06Ful} models sequential decision-making problems and allows for a more general definition of the agent's reward, which enables to express the agent's preferences more effectively.

Using the MDP notation, the behavior of agent-system pair can be modelled as follows.
\begin{definition}[Closed-loop model] \label{def:closed_loop_descr}
The behavior of the closed-loop formed of the agent-system pair up to time $t\in\mathbb{N}$
is described by a joint probability function $p(s_t,a_t,s_{t-1},\dots,s_1,a_1,s_0)$, where
$s_\tau\in\S{S}$, $0\leq\tau\leq t$, are states of the system and $a_\tau\in\S{A}$, $1\leq\tau\leq t$ denote actions of the agent.
\end{definition}

By applying the Markov property \cite{Put:94} and the chain rule for probabilities, the closed-loop model can be written in the form
\begin{equation}\label{eq:closed-loop_beh}
 \begin{aligned}
  p(s_t,&a_t,s_{t-1},\dots,s_1,a_1,s_0)  \\
  & {=}\: \prod_{\tau=1}^{t} p(s_\tau|a_\tau,s_{\tau-1})p_\tau(a_\tau|s_{\tau-1})p(s_0),
 \end{aligned}
\end{equation}
where the first factor $p(s_\tau|a_\tau,s_{\tau-1})$ is \emph{the transition model}, $p_\tau(a_\tau|s_{\tau-1})$ is \emph{the decision rule} at decision epoch $\tau$, and $p(s_0)$ represents the prior distribution of the initial state.

Instead of defining a reward function like in the MDP problem formulation, the agent's preferences over possible states and actions are expressed via an ideal closed-loop model.
\begin{definition}[Ideal closed-loop model\footnote{For brevity the term ideal model is sometimes used}]\label{def:ideal_closed-loop_beh}
A targeted behavior of the agent-system loop up to time $t\in\mathbb{N}$ is described by a joint probability function $\Ip(s_t,a_t,s_{t-1},\dots,s_1,a_1,s_0)$, $s_\tau\in\S{S}$ for $0\leq\tau\leq t$ and $a_\tau\in\S{A}$ for $1\leq\tau\leq t$.
\end{definition}
When factorising the ideal closed-loop model in a way similar to (\ref{eq:closed-loop_beh}), the first factor
describes targeted dynamics of the system and the second factor reflects possible preferences among possible actions.
Definition \ref{def:closed_loop_descr} and Definition \ref{def:ideal_closed-loop_beh} allow to formulate the underling DM problem via minimization of the Kullback-Leibler divergence between the actual closed-loop model (Definition~\ref{eq:closed-loop_beh}) and the desired closed-loop model (Definition~\ref{def:ideal_closed-loop_beh}). In other words, the optimal decision policy should make the closed-loop behavior as close as possible to the targeted one.
This is an essence of FPD \cite{Kar:96} and the FPD optimal policy can be defined as follows.
\begin{definition}[Optimal FPD decision policy]\label{def:FPD_opt_dec}
  An optimal FPD decision policy is defined as
  \begin{displaymath}
   \begin{aligned}
    \pi_{FPD}^{opt} = \underset{\big\{ p_t(a_t|s_{t-1})\big\}_{t=1}^{H}}{\arg\min} \pmb{D}\Big(&p(s_H,a_H,\dots,s_1,a_1,s_0) \\ & \big|\big| \Ip(s_H,a_H,\dots,s_1,a_1,s_0)\Big),
     \end{aligned}
  \end{displaymath}
  where $p_t(a_t|s_{t-1})$ a DM rule at time $t$, $H\in\mathbb{N}$ is optimization horizon, $s_\tau\in\S{S}$, for $0\leq\tau\leq H$, $a_\tau\in\S{A}$, for $1\leq\tau\leq H$, and $\pmb{D}(\cdot||\cdot)$ is the Kullback-Leibler divergence.
\end{definition}
The solution to FPD gives the following theorem.
\begin{theorem}[Solution to FPD]\label{th:solution_FPD}
The explicit optimal FPD policy minimizing the KL divergence (see Definition \ref{def:FPD_opt_dec}) is constructed using the following equations
    \begin{align*}
    \phantom{\op_t(a_t|s_{t-1})}
    &\begin{aligned}
    \mathllap{\op_t(a_t|s_{t-1})} = &{}\:\Ip_t(a_t|s_{t-1}) \\ &{\times}\:\frac{\exp\big(-\alpha(a_t,s_{t-1})-\beta(a_t,s_{t-1})\big)}{\gamma(s_{t-1})}
    \end{aligned}\\
    &\mathllap{\alpha(a_t,s_{t-1})} = \underset{s_t\in\S{S}}{\sum} p(s_t|a_t,s_{t-1})\ln\frac{p(s_t|a_t,s_{t-1})}{\Ip(s_t|a_t,s_{t-1})} \\
    &\mathllap{\beta(a_t,s_{t-1})} = -\underset{s_t\in\S{S}}{\sum}\ln(\gamma(s_{t}))p(s_t|a_t,s_{t-1}) \\
    &\begin{aligned}
    \mathllap{\gamma(s_{t-1})} = & \underset{a_t\in\S{A}}{\sum}\Ip_t(a_t|s_{t-1}) \\
    &{\times}\:\exp\big(-\alpha(a_t,s_{t-1})-\beta(a_t,s_{t-1})\big)
    \end{aligned}\\
    &\mathllap{\gamma(s_{H})} = 1
    \end{align*}
  for all $t\in\{1,\dots,H\}$, where $H\in\mathbb{N}$ is a horizon of optimization.
\end{theorem}
\begin{IEEEproof}
  See \cite{Kar:96}.
\end{IEEEproof}
\vspace{2mm}
\textbf{Relation between FPD and MDP:} FPD formulation is more general and an MDP problem can be formulated and solved as FPD problem with
\begin{equation} \label{eq:closed-loop_factoris}
  r(s_t,a_t,s_{t-1}) = -\ln\frac{p(s_t,a_t|s_{t-1})}{\Ip(s_t,a_t|s_{t-1})}.
\end{equation}
The relation (\ref{eq:closed-loop_factoris}) results from a direct application of the Kullback-Leibler divergence and  Definition \ref{def:FPD_opt_dec}.

\section{Similarity-Based Transfer Learning}\label{sec:TL}

In this section we present an approach to learning a DM policy from the a past data.
The approach is based on the probabilistic modeling used within FPD methodology and newly introduced similarity function.

The agent interacts with the system and aims to find an optimal DM policy that ensures reaching the targeted DM preferences.
Let us also suppose that there are data describing the \emph{past} closed-loop behaviour formed of the same system and generally different agent. Such data may be obtained from the experts (demonstration or training data) or describe solution of other DM tasks solved on the same system. The past policies are assumed to be consistent with some unknown ideal model, though not necessary optimal. The past ideal closed-loop models can significantly differ from the current one thus the past data should not match the current DM objective.
We are interested in learning the optimal DM policy from these data, i.e. in
transferring the best experience gained on the system to the present DM task.

\subsection{Solution Concept and Similarity Function}\label{sec:Solution_Concept}
 Let the agent sequentially interact with the system. The agent's DM preferences are expressed via ideal model $\Ip$ (see Definition \ref{def:ideal_closed-loop_beh}). We need to find an optimal sequence of DM rules ensuring reaching this ideal. Consider past data $\{(s_{\tau},a_\tau,s_{\tau-1})\}_{\tau=1}^{t-1}$  describing the previous, already completed, DM task. Actions $a_{\tau}, \tau=1,...,t-1$ are optimal with respect to past (and unknown) ideal model ${}^I\tilde{p}$ that can generally be different from  $\Ip$.

The proposed approach learns a sequence of DM rules for the current DM task from the past data available.
The key idea uses the fact that behavior of any system is substantially determined by fixed dependencies (for instance given by the first principles) that are independent of states and actions. Besides, data communicate indirect information about decision patterns\footnote{i.e. dependence "system state-corresponding action-next state"} applied in the past. Once the experience was collected on the same system, we can learn DM rules that suit to the current DM objectives.
To distinguish relevant experience we should be able to measure "degree of matching" the past data to the current DM objectives. To evaluate that, we introduce the term of similarity quantifying the extent to which the past behavior fits the current DM aim.

\begin{definition}[Similarity]\label{def:similarity}
  Let $\{(s_{\tau},a_\tau,s_{\tau-1})\}_{\tau=1}^{t-1}$ be a set of observations of a completed decision-making task. We define the \emph{similarity} between the current decision problem with the ideal model $\Ip$ and a past problem from decision epoch $\tau$ as
  \begin{equation}\label{eq:similarity}
    \sigma_\tau = \Ip(s_\tau,a_\tau|s_{\tau-1})\in[0,1],
  \end{equation}
  where $(s_{\tau},a_\tau,s_{\tau-1})$ is an observation of one decision and state transition, and $0<\tau< t$.
\end{definition}
The definition of similarity, has a clear and intuitive meaning. Whenever past data $(s_\tau,a_\tau,s_{\tau-1})$ bring high values of the current ideal model $\Ip$\footnote{or by other words the likelihood is high}, see Definition~\ref{def:ideal_closed-loop_beh}, past transition $(s_{\tau-1}, a_{\tau})\rightarrow  s_\tau$ , is close to the targeted behavior in the current DM problem.
The value of the similarity is small whenever a past decision pattern: i) simulates state transition that does not fully match the current DM preferences (expressed by ideal model $\Ip$), ii) is considered disadvantageous for the current DM preferences. If the past system transition is desirable with regard to the current DM problem, the similarity is high.

The following definition of similarity is almost identical to Definition \ref{def:similarity}, except the values are normalized.
\begin{definition}[Normalized similarity]\label{def:similarity_norm}
  We will consider the set $\{(s_{\tau},a_\tau,s_{\tau-1})\}_{\tau=1}^{t-1}$ be a set of observations of a completed decision-making task. The \emph{normalized similarity} between the current decision problem with the ideal model $\Ip$ and a past problem from decision epoch $\tau$, $0<\tau< t$, is defined as
  \begin{equation}\label{eq:similarity_norm}
  \begin{gathered}
    \sigma_\tau = \frac{\Ip(s_\tau,a_\tau|s_{\tau-1})}{\sigma_{max}}\in[0,1], \ \text{where}\\
    \sigma_{max} = \underset{s_t,s_{t-1}\in\S{S}, a_t\in\S{A}}{\max} \Ip(s_t,a_t|s_{t-1}).
    \end{gathered}
  \end{equation}
\end{definition}

Introducing a normalized version of the similarity is important because similarity equals to an ideal likelihood of past data, so it's maximum possible value is not in the interval $[0,1]$. Therefore to judge whether the obtained value of similarity is high enough we them to normalise.

Note that Definition \ref{def:similarity} and Definition \ref{def:similarity_norm} can be used in case the past data is the only information available, i.e. no information about the past DM preferences is available. It is clear that if past ideal models are known, the similarity can be measured via any divergence measure on the space of probability distributions.

\subsection{Bayes Estimation of the Decision Policy}\label{sec:Bayes_learning}

This section introduces Bayesian approach \cite{Pet:81} that guides the optimal DM rule selection based on the past data. Throughout this section, $(s',a)\rightarrow s$ denotes an arbitrary system state transition $(s_{\tau-1},a_\tau) \rightarrow s_\tau$, where $\tau\in\S{T}$.

Consider a DM task characterized by ideal model $\Ip$ and past data collected on the same system though for a different DM task. The data consists of a sequence of state transitions $d_{t-1}=\{(s_{\tau},a_\tau,s_{\tau-1})\}_{\tau=1}^{t-1}$ and our goal is to infer the targeted DM rule using $d_{t-1}$.

Generally the unknown closed-loop model $p(s_t,a_t|s_{t-1})$, which implicitly contains DM rule $p(a_t|s_{t-1})$ at time $t$, can be parameterized as $p(s_t,a_t|s_{t-1},\theta)$, where $\theta\in\mathbf{\Theta}$ in an unknown finite-dimensional parameter and $\mathbf{\Theta}$ is a continuous parameter space. We define the parameter space as
\begin{displaymath}
\begin{aligned}
  \mathbf{\Theta} = \bigg\{\theta_{s,a|s'}\big| s,s'\in\S{S}&,a\in\S{S}, \theta_{s,a|s'}\in[0,1],\\
  &\sum_{s\in\S{S},a\in\S{A}}\theta_{s,a|s'} = 1, \forall s'\in\S{S}\bigg\},
  \end{aligned}
\end{displaymath}
and the parametrization as $\theta_{s_t,a_t|s_{t-1}} = p(s_t,a_t|s_{t-1},\theta)$.

The closed-loop behavior based on the observed data at time $t$ is then described using marginalization and the chain rule as
\begin{equation}\label{eq:closed-loop_beh_data}
\hat{p}(s_t,a_t|d_{t-1}) = \int_\Theta p(s_t,a_t|d_{t-1},\theta)p(\theta|d_{t-1}) d\theta.
\end{equation}
The second factor $p(\theta|d_{t-1})$ in (\ref{eq:closed-loop_beh_data}) is a distribution expressing our beliefs about the unknown parameter based on $d_{t-1}$.

Using the available data $d_{t-1}$ we can write the posterior distribution of the parameter at decision epoch $t\in\mathbb{N}$ via the weighted Bayes' rule \cite{Kar:14Laz}:
 \begin{equation}\label{eq:weighted_bayes}
  p(\theta|d_{t-1})= \frac{\prod_{\tau=1}^{t-1}p(s_\tau,a_\tau|s_{\tau-1},\theta)^{\omega_\tau}p(\theta|s_0)}
  {\int_\Theta\prod_{\tau=1}^{t-1}p(s_\tau,a_\tau|s_{\tau-1},\theta)^{\omega_\tau}p(\theta|s_0)d\theta}.
\end{equation}
In (\ref{eq:weighted_bayes}),  $\omega_\tau=\sigma_\tau$ are values of the similarity, (\ref{eq:similarity_norm}) that numerically express how data $d_{t-1}$ fit the current ideal model $\Ip$.
New system transition, $s_{t-1}\rightarrow s_{t}$, enriches data with the tuple $(s_t,a_t,s_{t-1})$, and the posterior distribution can be updated to $p(\theta|d_{t})$ via (\ref{eq:weighted_bayes}). We simplified the formula (\ref{eq:weighted_bayes}) using the Markov property (\ref{eq:closed-loop_beh}) stating that the system state transition depends on the last state and action only.

It is assumed that initial state $s_0$ does not change the prior beliefs about parameters of the closed-loop model, i.e. $p(\theta|s_0) = p(\theta)$. This assumption is justified by considering the initial state as an initial condition not dependent on the parameter \cite{Pet:81}.

Additionally, we assume that $\theta_{s_t,a_t|s_{t-1}} = p(s_t,a_t|s_{t-1},\theta)$ follows multinomial distribution and prior $p(\theta)$  is a product of Dirichlet distributions\footnote{Dirichlet distribution as prior is a common choice in Bayesian theory. It simplifies the computation of the posterior distribution because the prior and the posterior distributions are conjugate for multinomial distribution sampling \cite{Fer74}}:
\begin{equation}\label{eq:dirichlet_prior}
  p(\theta) =
  \prod_{s'\in\S{S}}\text{Dir}\Big(\theta_{\cdot,\cdot|s'},\nu_0^{\cdot,\cdot|s'}\Big),
\end{equation}
where $\theta_{\cdot,\cdot|s'}$ is a vector of parameters, and $\nu_0^{\cdot,\cdot|s'}$ is a vector of values $\nu_0^{s,a|s'}>0$, $s\in\S{S}$, $a\in\S{A}$. Then the posterior obtained using the weighted Bayes rule (\ref{eq:weighted_bayes}) has the form
\begin{equation}\label{eq:posterior}
  p(\theta|d_{t-1}) \propto \prod_{s'\in\S{S}}\text{Dir}\Big(\theta_{\cdot,\cdot|s'},V_{t-1}^{\cdot,\cdot|s'}\Big)
\end{equation}
with concentration parameters defined recursively $\forall 1\leq\tau \leq t-1$
\begin{align*}
  \phantom{V_\tau^{s,a|s'}}
  & \begin{aligned}
  \mathllap{V_\tau^{s,a|s'}} = \omega_\tau\delta(s,s_\tau)\delta(a,a_\tau)\delta(s',s_{\tau-1}) + V_{\tau-1}^{s,a|s'},
  \end{aligned}\\
  &\mathllap{V_0^{s,a|s'}} = \nu_0^{s,a|s'}.
\end{align*}
where $\delta(\cdot,\cdot)$ is the Kronecker delta function and $(s_\tau,a_\tau,s_{\tau-1})\in d_{t-1}$, $1\leq\tau\leq t-1$, are \emph{observed} realizations of states and actions. These realizations describe the possible closed-loop transitions. These learned parameters essentially represent the number of transitions $(s',a)\rightarrow s$, that were observed in the past, weighted by "usefulness" of a particular transition for the current DM problem.

The deduced form of the posterior distribution (\ref{eq:posterior}) is then used to derive the learned optimal DM rule
\begin{equation}\label{eq:estimated_opt_dec_rule_theory}
\hat{p}(a_t|s_{t-1}) = \sum_{s_t\in\S{S}}\int_\Theta p(s_t,a_t|s_{t-1},\theta)p(\theta|d_{t-1}) d\theta.
\end{equation}
After the computing using the definition of the Beta function that appears as a normalizing constant in the Dirichlet distribution, and utilizing properties of the Gamma function, (\ref{eq:estimated_opt_dec_rule_theory}) can be rewritten as
\begin{equation}\label{eq:estimated_dec_rule}
\begin{aligned}
  {}^{opt}&\hat{p}(a_t|s_{t-1}) = \\
  &{=}\:\frac{\sum_{\tau=1}^{t-1}\omega_\tau\delta(a_t,a_\tau)\delta(s_{t-1},s_{\tau-1})+\sum_{s\in\S{S}}\nu_0^{s,a_t|s_{t-1}}}
  {\sum_{\tau=1}^{t-1}\omega_\tau\delta(s_{t-1},s_{\tau-1})+\sum_{s\in\S{S}}\sum_{a\in\S{A}}\nu_0^{s,a|s_{t-1}}}.
\end{aligned}
\end{equation}

The formula (\ref{eq:estimated_dec_rule}) gives an optimal DM rule that was learned from the past history. The learned rule comprises the best past experience which can be useful for the current DM objective.  Note that this rule, though called optimal, is an approximation of the unknown optimal rule.

\subsection{Exploration}\label{sec:exploration}

The approach proposed above exploits all available information about the closed-loop behavior the best experience available. However, past data can be i) incomplete; ii) obtained for DM preferences significantly differing from current objectives (defined by the ideal model $\Ip$). Then an exploration ability should be added as it helps to gather more information about the system.

A computationally inexpensive exploration strategy is the $\epsilon$-greedy explorative strategy. It was introduced in \cite{Wat89} as a strategy solving the multi-armed bandit problem. It chooses the currently optimal action (i.e. the optimal decision rule \eqref{eq:estimated_dec_rule}) with probability 1-$\epsilon$ and a random action with probability $\epsilon$, $\epsilon\in[0,1]$.

To prevent unnecessary over-exploration, an $\epsilon$-greedy exploration technique is applied whenever there is a lack of  data, which is visible from the mean value of $m\in\mathbf{N}$ last computed similarities. If the mean is lower than a given threshold $q\in[0,1]$, $\epsilon$-greedy exploration is activated. Otherwise, the learned optimal decision rule \eqref{eq:estimated_dec_rule} is applied directly.

The resulting algorithm of finding the optimal decision rule with incorporated exploration and using the normalized version of the similarity (\ref{eq:similarity_norm}) is shown in Algorithm \ref{alg:alg_exploration}.

\begin{algorithm}
\caption{Transfer learning of an optimal decision rule with exploration}\label{alg:alg_exploration}
  \begin{algorithmic}
 \Require data $d_{t-1}=\left\{(s_\tau,a_{\tau},s_{\tau-1})\right\}_{\tau=1}^{t-1}$, ideal model $\Ip$
 \For{$\tau = 1,\dots,t-1$}
    \State Compute weight $\omega_\tau \equiv \sigma_\tau = \frac{\Ip(s_\tau,a_\tau|s_{\tau-1})}{\sigma_{max}}$;
 \EndFor
 \While{$t\leq N$}
    \State Learn the optimal decision rule ${}^{opt}\hat{p}(a_t|s_{t-1})$ \eqref{eq:estimated_dec_rule};
    \If{$\frac{1}{m}\sum_{\tau=t-m}^{t-1}\omega_\tau < q$}
        \State Generate $\xi_t$ from a uniform distribution on the
        \State interval [0,1];
        \If{$\xi_t < \epsilon$}
            \State Use the uniform decision rule $p_t(a_t|s_{t-1}) = \frac{1}{|\S{A}|}$;
        \Else
            \State Use the learned decision rule (\ref{eq:estimated_dec_rule});
        \EndIf
    \Else
        \State Use the learned decision rule (\ref{eq:estimated_dec_rule});
    \EndIf
    \State Observe new state transition $(s_{t-1},a_t)\rightarrow s_t$;
    \State Calculate new weight $\omega_t \equiv \sigma_t = \frac{\Ip(s_t,a_t|s_{t-1})}{\sigma_{max}}$;
    \State $t = t+1$;
 \EndWhile
 \end{algorithmic}
\end{algorithm}

\section{Illustrative experiments} \label{sec:Illustrative_experiments}

The proposed approach was demonstrated and verified through a series of simulated experiments.
Each experiment was repeated 100 times.

\noindent
\textbf{General setting. } We consider a discrete system with state space $\S{S}=\{s^1,s^2,s^3\}$.  The action space contains four actions, $\S{A}=\{a^1,a^2,a^3,a^4\}$.
The particular coefficients of the transition model, $p(s_t|a_t,s_{t-1})$, were generated randomly, so the system dynamics, was different each time.
Initial state $s_0$ was also chosen randomly with respect to the uniform distribution.
The overall experiment tasks were as follows:
\begin{itemize}
  \item generate data for different DM tasks (determined by different DM preferences), i.e. imitate past experience
  \item set a new DM task characterised buy new DM preferences
  \item learn the optimal DM policy for a new DM task by using the proposed transfer learning
  \item apply the DM policy learned and compare the obtained close-loop performance.
\end{itemize}

\noindent
\textbf{How were the past data generated?} To simplify further comparison and verification of the proposed approach, the FPD settings were used for past data generating. The following experiment was performed.
The past DM objectives were set and expressed via ideal model (see Definition~\ref{def:ideal_closed-loop_beh}). Then optimal FPD decision policy (Theorem~\ref{th:solution_FPD}) was computed and applied to the system. The optimal policy was computed for the completely known transition model (no mismodeling).
The horizon of the policy optimization was set to $H=10$ decision epochs. The resulting closed-loop behavior was observed over $k=60$ decision epochs, so the past data available were $d_{60} = \left\{(s_\tau,a_\tau,s_{\tau-1})\right\}_{\tau = 1}^{60}$. These data were further used for transfer learning.

\noindent
\textbf{What were DM preferences of the past DM tasks?}
Three different ideal transition models were used during the generation of the demonstration data $d_{60}$. For all of them, the ideal decision rule was uniform, i.e. no preference over actions existed. The first ideal model, labeled as ${}^I\tilde{p}_1$, favored state $s^1$ and was defined as
\begin{equation}\label{eq:past_model_1}
  \begin{gathered}
  {}^I\tilde{p}_1(s_t = s^1|a_t,s_{t-1}) = 0.99998, \\
  {}^I\tilde{p}_1(s_t\neq s^1|a_t,s_{t-1}) = 0.00001,
  \end{gathered}
\end{equation}
for all $a_t\in\S{A}$, $s_{t-1}\in\S{S}$. The second ideal model, ${}^I\tilde{p}_{1,2}$, reflected equal preference for $s^1$ and $s^2$. For all $a_t\in\S{A}$ and all $s_{t-1}\in\S{S}$ it was defined as follows
\begin{equation}\label{eq:past_model_12}
  \begin{gathered}
  {}^I\tilde{p}_{1,2}(s_t = s^1|a_t,s_{t-1}) = 0.499995, \\
  {}^I\tilde{p}_{1,2}(s_t = s^2|a_t,s_{t-1}) = 0.499995, \\
  {}^I\tilde{p}_{1,2}(s_t = s^3|a_t,s_{t-1}) = 0.00001.
  \end{gathered}
\end{equation}
The third ideal model ${}^I\tilde{p}_3$ favored state $s^3$ only:
\begin{equation}\label{eq:past_model_3}
  \begin{gathered}
  {}^I\tilde{p}_3(s_t = s^3|a_t,s_{t-1}) = 0.99998, \\
  {}^I\tilde{p}_3(s_t\neq s^3|a_t,s_{t-1}) = 0.00001,
  \end{gathered}
\end{equation}
for all $a_t\in\S{A}$ and for all $s_{t-1}\in\S{S}$.

There were no special preferences on actions. Thus the ideal decision rule was a uniform probability function: $\Ip(a_t|s_{t-1})=\frac{1}{|\S{A}|} = 0.25$, for all $a_t\in\S{A}$, $s_{t-1}\in\S{S}$. The agent's ideal transition model $\Ip$ was the same as ${}^I\tilde{p}_1$ (\ref{eq:past_model_1}), focused on reaching state $s^1$. It was
\begin{equation}\label{eq:current_ideal_model_1}
  \begin{gathered}
  \Ip(s_t = s^1|a_t,s_{t-1}) = 0.99998, \\
  \Ip(s_t\neq s^1|a_t,s_{t-1}) = 0.00001,
  \end{gathered}
\end{equation}
for all $a_t\in\S{A}$, $s_{t-1}\in\S{S}$.

\noindent
\textbf{How the results were compared?}
With the past data $d_{60}$ collected, a optimal decision policy with respect to ideal $\Ip$ for $h=100$ was searched. Normalized version of the similarity (\ref{eq:similarity_norm}) was used to weight the past observations.

The verify the proposed approach, the optimal DM policy for the current DM task was searched via different algorithms (names correspond to the notations used in Fig.~\ref{fig:103}-~Fig.~\ref{fig:dif}):
\begin{description}
  \item[Rand] \hspace{3mm}- random policy;
  \item[$\text{TL}$] \hspace{3mm}- the proposed similarity-based transfer learning an optimal policy without exploration, (Section \ref{sec:TL})
  \item[$\text{TL}_{\text{explore}}$] \hspace{3mm}- the proposed similarity-based transfer learning an optimal policy with exploration strategy, Section~\ref{sec:exploration}.
  \item[$\text{FPD}_{\text{learn}}$] \hspace{3mm}- FPD method Section~\ref{th:solution_FPD} when the transition model is unknown and learned on-line
  \item[FPD] \hspace{3mm}- FPD method using the complete knowledge of the transition model.
\end{description}
 The closed-loop behaviors corresponding to the different methods of policy generation were then compared based on the closed-loop performance.

The performance of the TL method was measured by \emph{gain}, which was defined as the overall number of occurrences of state $s^1$. Prior distribution parameters (\ref{eq:dirichlet_prior}) were chosen so that they were all equal to
\begin{equation*}
  \nu_0 = \frac{1}{|\S{S}|}\underset{\substack{s_t,s_{t-1}\in\S{S}\\a_t\in\S{A}}}{\min}\Ip(s_t,a_t|s_{t-1}),
\end{equation*}
which suggests no prior information about the parameter of the closed-loop model.

The seed for reproducibility of results was set to $10$. The methods and experiments were implemented in Matlab R2016b\textregistered. Boxplot figures were generated using Alternative box plot function for Matlab from the IoSR Matlab Toolbox \cite{Hum:16}.

\subsection{Comparison of the TL and the FPD methods}

\begin{figure}[t!]
    \includegraphics[width=\linewidth]{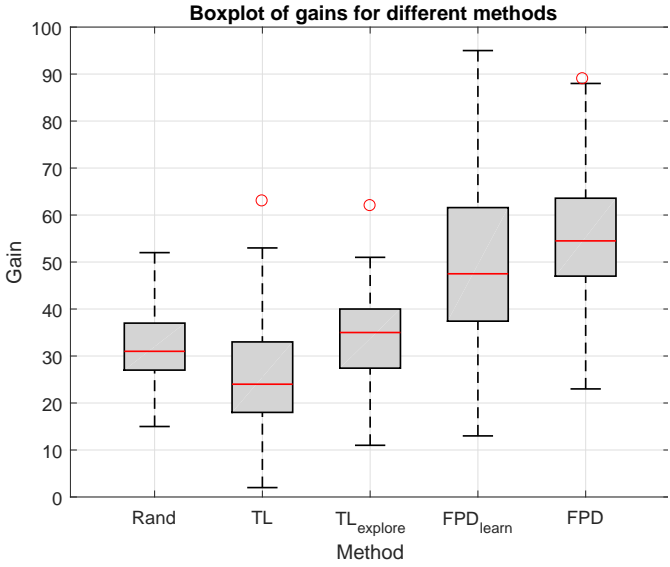}
    \caption{Boxplot of gains, data gathered with ideal model ${}^I\tilde{p}_3$. Rand - random policy, TL - TL method, $\text{TL}_{\text{explore}}$ - TL method with exploration, $\text{FPD}_{\text{learn}}$ - learning FPD method, FPD - FPD method with complete knowledge.} \label{fig:103}
\end{figure}

The TL method was used either without any exploration (\ref{eq:estimated_dec_rule}), or with adjusted exploration strategy (see Algorithm \ref{alg:alg_exploration}). Then the exploration rate was set to $\epsilon=0.3$, the threshold of low average similarity was $q=0.4$, and the number of previous similarities to be averaged was $m=10$.

The FPD method (Theorem \ref{th:solution_FPD}) was employed either with complete knowledge of the transition model $p(s_t|a_t,s_{t-1})$, or without any prior knowledge of the model. In the latter case, Bayesian estimation was applied to learn the transition model using the same set of observations $d_{60}$ as those available for the TL method. The case with complete knowledge of the transition model represents a boundary situation because it is not common in real-life applications and served to comparison only.
FPD policy was optimized over a horizon of $H=10$ decision epochs in both cases.

The two methods were also compared to a \emph{random policy}, that is a policy that chooses actions randomly at each decision epoch and is defined for all $a_t\in\S{A}$ and all $s_{t-1}\in\S{S}$ as
\begin{equation}\label{eq:random_strategy}
  p(a_t|s_{t-1}) = \frac{1}{|\S{A}|}.
\end{equation}

\begin{figure}[t!]
    \includegraphics[width=\linewidth]{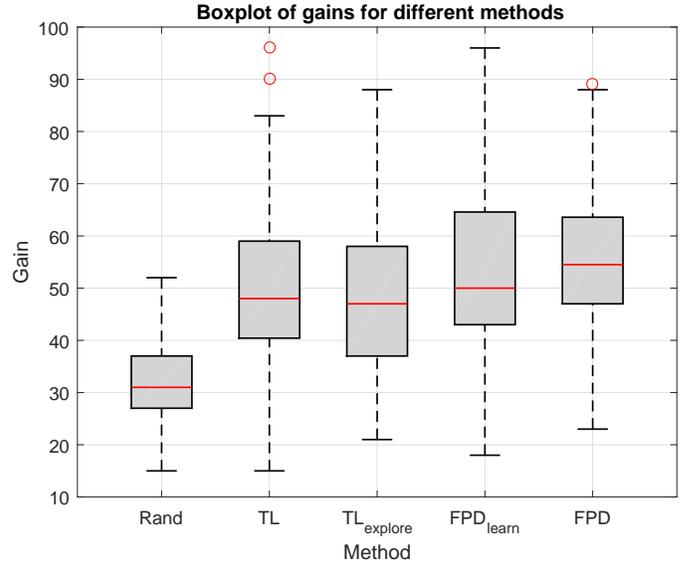}
    \caption{Boxplot of gains, data gathered with ideal model ${}^I\tilde{p}_{1,2}$. Rand - random policy, TL - TL method, $\text{TL}_{\text{explore}}$ - TL method with exploration, $\text{FPD}_{\text{learn}}$ - learning FPD method, FPD - FPD method with complete knowledge.} \label{fig:112}
\end{figure}

\begin{figure}[t!]
    \includegraphics[width=\linewidth]{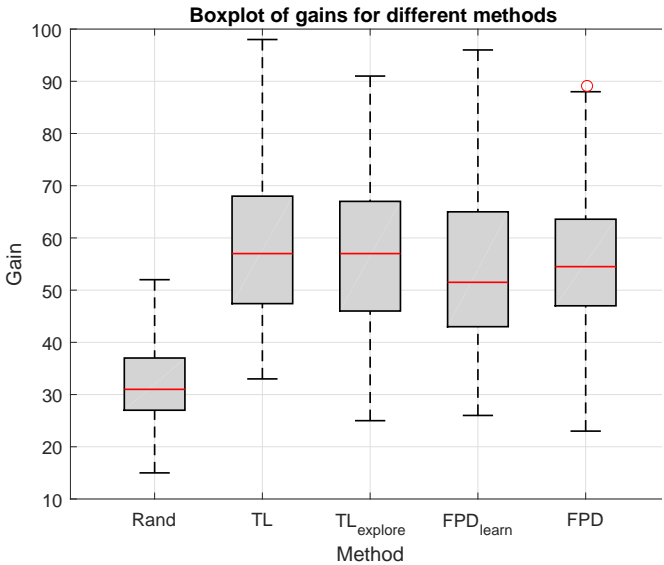}
    \caption{Boxplot of gains of different methods, data gathered using ideal transition model ${}^I\tilde{p}_1$. Rand - random policy, TL - TL method, $\text{TL}_{\text{explore}}$ - TL method with exploration, $\text{FPD}_{\text{learn}}$ - learning FPD method, FPD - FPD method with complete knowledge.} \label{fig:101}
    \end{figure}

Fig.~\ref{fig:103} shows a boxplot representing results of a method comparison where past data $d_{60}$ were collected using ideal transition model ${}^I\tilde{p}_3$ (\ref{eq:past_model_3}), so with completely different DM preferences than the current ideal model, $\Ip$ (\ref{eq:current_ideal_model_1}). Fig.~ \ref{fig:112} illustrates results of the experiment for the past ideal model, ${}^I\tilde{p}_{1,2}$, (\ref{eq:past_model_12}) that expresses DM preferences that partly overlap with the current ones, expressed via $\Ip$. Fig.~ \ref{fig:101} represents gains of the compared methods when pas data $d_{60}$ generated with ${}^I\tilde{p}_1$ (\ref{eq:past_model_1}), i.e. DM preferences of the past and current tasks coincide. 

Fig.~ \ref{fig:103} illustrates that when there is no overlap of past and present objectives, the TL performs worse even than the random policy. When TL with exploration was used, the gains rose slightly above the random policy gains. However, they were still considerably worse than the FPD ones. As shown in Fig.~\ref{fig:112}, the results improved greatly when the decison policy had been learned from past data more relevant to the current DM preferences. The performance of the TL is nearly equal to that of the FPD with completely known transition model. Finally, as can be seen in Fig.~\ref{fig:101}, the TL method outperforms the FPD method in the conditions of data matching current objectives. Note that the exploration strategy worsened the results only slightly when past data were appropriate.

\begin{figure}[ht]
    \includegraphics[width=\linewidth]{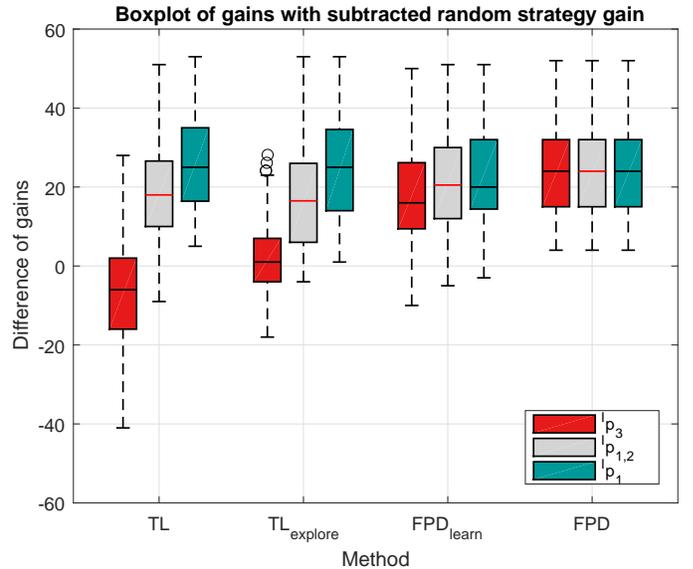}
    \caption{Boxplot of gains of different methods with gain of random policy subtracted, data gathered with three different ideal models ${}^I\tilde{p}$. TL - TL method, $\text{TL}_{\text{explore}}$ - TL method with exploration, $\text{FPD}_{\text{learn}}$ - learning FPD method, FPD - FPD method with complete knowledge.} \label{fig:dif}
\end{figure}

Results of the same experiments as in Fig.~s\ref{fig:103}, \ref{fig:112} and \ref{fig:101} are shown in Fig.~\ref{fig:dif}, where gains of the random policy were subtracted from gains of other methods. The transition model parameters were different for each simulation so the difficulty of obtaining the desired states varied. Fig.~\ref{fig:dif} depicts success of each DM policy compare to the random policy depending on the quality of the past data used.  
Naturally the results of the FPD method with complete knowledge of the transition model were the same for all three types of data because the method did not need to use the data to estimate the transition model.

\subsection{Computational complexity}

An important aspect of an algorithm is its computational complexity. The complexity of determining one decision rule was estimated using the "big $O$" notation \cite{Ull:94}, which indicates asymptotic number of operations. It can be considered as an upper bound of the complexity. Estimating the optimal decision rule using the TL method with exploration and with normalized similarity, see Algorithm \ref{alg:alg_exploration}, takes asymptotically $O(\max(k,|\S{S}|^2\cdot|\S{A}|))$ operations, where $k$ is the number of past observations available (length od the data), $|\S{S}|$ is the number of states and $|\S{A}|$ is the number of actions. When determining the decision rule, the first step is computing the similarities using the data of length $k$. The similarities are then normalized, so a normalizing constant has to be found as a maximum value of the ideal model $\Ip(s_t,a_t|s_{t-1})$, which has the dimensions of $|\S{S}|\cdot|\S{A}|\cdot|\S{S}|$. Lastly, the decision rule is learnt using the computed $k$ similarities. Multiplicative and additive constants are omitted because the "big $O$" symbol describes the asymptotic long-term growth of the number of operations.

Computing the optimal decision policy with FPD learning method takes $O(\max(k,H\cdot|\S{S}|^2\cdot|\S{A}|))$ operations, where $H$ is the horizon of policy optimization. First, the unknown transition model has to be estimated using the $k$ observations, then the optimal decision rule is computed (Theorem \ref{th:solution_FPD}) over the horizon $H$. In our experiment $H$ was set to 10, so it can be considered as a constant and omitted. Then both methods have the same theoretical asymptotic complexity $O(\max(k,|\S{S}|^2\cdot|\S{A}|))$.

In practice, the omitted coefficients and constants as well as other factors are important for the true computational time. That is why it is necessary to carry out experiments measuring the real time complexity of both algorithms. An experiment was conducted comparing the CPU time required for computing the decision rule using the TL method with exploration and the learning FPD method. The CPU time was determined using the Matlab\textregistered \ in-built \emph{timeit} function. It runs a specified function several times and returns the median of the elapsed times. The CPU time depends on the computer used, thus all results should be perceived as an illustration of the expected behavior. The computer used to provide the results presented here was SAMSUNG 900X3C, 2.00 GHz Intel Core i7 with 4GB RAM.

\begin{figure}[t!]
    \includegraphics[width=\linewidth]{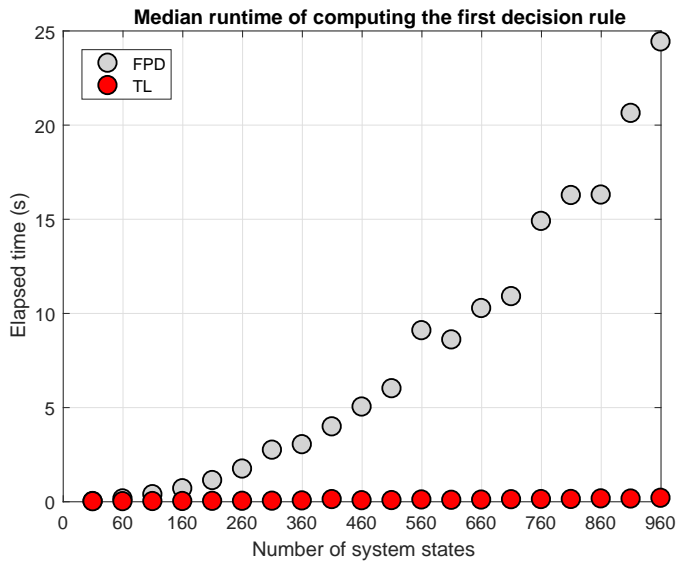}
    \caption{Median CPU time required to determine the first decision rule after obtaining the data $d_k$ for growing number of states using the FPD learning and the TL method with exploration.} \label{fig:time_ns_first}
\end{figure}

In Fig.~ \ref{fig:time_ns_first}, the median time complexity of computing the first decision rule with changing number of states $|\S{S}|$ is shown. The number of observations was fixed at $k = 30$, the number of actions was fixed at $|\S{A}| = 4$. It can be noted that the elapsed time using the FPD method increases much faster for growing $|\S{S}|$ than the elapsed time using the TL method. Even though the theoretical asymptotic complexity is the same for both, the real time complexity is significantly smaller for high number of states using the TL method. The true order of complexity of the TL is possibly lower than the true order of complexity of the FPD.

\section{Concluding remarks}\label{sec:Conclusions}

The sequential decision making was considered. The paper proposes learning an optimal decision policy using the experience gained during solving other DM tasks on the same system.
 The approach related to a class of approaches like imitation learning, apprenticeship learning while  uses the whole past experience available irrespectively of i)the past DM objectives; ii)quality of applied policies, and iii) the resulting overall success. The key features of the proposed solution are:
\begin{itemize}
  \item Useful experience occurred in the past will be amplified and transferred to a new decision policy.
  \item Useless (and even harmful) experience will not be neglected but transferred with much smaller weights. This allows to make learned decision policy "aware" of possible bad consequences without experiencing them in reality.
  \item Possible non-optimality of the past decision policies serve as a natural source of exploration.
  \item The proposed solution is robust to the errors that can be transferred from the past as the resulting policy comprises all kinds of past behaviours: successful and not.
  \item There is no need to use expert's demonstration data, the good experience can come from anywhere (data coming from dozens of non-experienced drivers may give rise the decision policy that overcomes an expert's policy).
\end{itemize}

Further research will consider: i) real-application experiments, ii) the possibility to construct and work with complex multi-dimensional DM preferences.

\bibliographystyle{IEEEtran}
\bibliography{IEEEabrv,Bib_conference} 

\begin{thebibliography}{10}
\providecommand{\url}[1]{#1}
\csname url@samestyle\endcsname
\providecommand{\newblock}{\relax}
\providecommand{\bibinfo}[2]{#2}
\providecommand{\BIBentrySTDinterwordspacing}{\spaceskip=0pt\relax}
\providecommand{\BIBentryALTinterwordstretchfactor}{4}
\providecommand{\BIBentryALTinterwordspacing}{\spaceskip=\fontdimen2\font plus
\BIBentryALTinterwordstretchfactor\fontdimen3\font minus
  \fontdimen4\font\relax}
\providecommand{\BIBforeignlanguage}[2]{{%
\expandafter\ifx\csname l@#1\endcsname\relax
\typeout{** WARNING: IEEEtran.bst: No hyphenation pattern has been}%
\typeout{** loaded for the language `#1'. Using the pattern for}%
\typeout{** the default language instead.}%
\else
\language=\csname l@#1\endcsname
\fi
#2}}
\providecommand{\BIBdecl}{\relax}
\BIBdecl

\bibitem{Wu_etal:19}
Y.-H. Wu, N.~Charoenphakdee, H.~Bao, V.~Tangkaratt, and M.~Sugiyama,
  ``Imitation learning from imperfect demonstration,'' in \emph{Proceedings of
  the 36th International Conference on Machine Learning}, ser. Proceedings of
  Machine Learning Research, vol.~97.\hskip 1em plus 0.5em minus 0.4em\relax
  Long Beach, California, USA: PMLR, 09--15 Jun 2019, pp. 6818--6827.

\bibitem{AbbNg:04}
P.~Abbeel and A.~Y. Ng, ``Apprenticeship learning via inverse reinforcement
  learning,'' in \emph{Proceedings of the Twenty-First International Conference
  on Machine Learning}, ser. ICML ’04.\hskip 1em plus 0.5em minus 0.4em\relax
  New York, NY, USA: Association for Computing Machinery, 2004.

\bibitem{Cha_etal:15}
K.-W. Chang, A.~Krishnamurthy, A.~Agarwal, H.~Daum\'{e}, and J.~Langford,
  ``Learning to search better than your teacher,'' in \emph{Proceedings of the
  32nd International Conference on International Conference on Machine Learning
  - Volume 37}, ser. ICML’15, 2015, p. 2058–2066.

\bibitem{Kau_etal19}
E.~{Kaufmann}, M.~{Gehrig}, P.~{Foehn}, R.~{Ranftl}, A.~{Dosovitskiy},
  V.~{Koltun}, and D.~{Scaramuzza}, ``Beauty and the beast: Optimal methods
  meet learning for drone racing,'' in \emph{2019 International Conference on
  Robotics and Automation (ICRA)}, 2019, pp. 690--696.

\bibitem{Lev_etal:09}
S.~Levine, C.~Theobalt, and V.~Koltun, ``Real-time prosody-driven synthesis of
  body language,'' \emph{ACM Trans. Graph.}, vol.~28, no.~5, 2009.

\bibitem{RosBag:2010}
S.~Ross and D.~Bagnell, ``Efficient reductions for imitation learning,'' in
  \emph{Proceedings of the Thirteenth International Conference on Artificial
  Intelligence and Statistics}, ser. Proceedings of Machine Learning Research,
  Y.~W. Teh and M.~Titterington, Eds., vol.~9, 2010, pp. 661--668.

\bibitem{Kar:06Ful}
M.~K\'{a}rn\'{y} and T.~Guy, ``Fully probabilistic control design,''
  \emph{Systems \& Control Letters}, vol.~55, no.~4, pp. 259--265, 2006.

\bibitem{Kar:20}
M.~K\'{a}rn\'{y}, ``Fully probabilistic design unifies and supports dynamic
  decision making under uncertainty,'' \emph{Information Sciences}, vol. 509,
  pp. 104 -- 118, 2020.

\bibitem{Put:94}
M.~Puterman, \emph{Markov Decission Processes}.\hskip 1em plus 0.5em minus
  0.4em\relax John Wiley \& Sons, Inc., 1994.

\bibitem{Kar:96}
M.~K\'{a}rn\'{y}, ``Towards fully probabilistic control design,''
  \emph{Automatica}, vol.~32, no.~12, pp. 1719--1722, 1996.

\bibitem{Kar:06Opt}
M.~K\'{a}rn\'{y}, J.~B\"{o}hm, T.~V. Guy, L.~Jirsa, I.~Nagy, P.~Nedoma, and
  L.~Tesa\v{r}, \emph{Optimized Bayesian dynamic advising}.\hskip 1em plus
  0.5em minus 0.4em\relax Springer London, 2006.

\bibitem{Pet:81}
V.~Peterka, ``Bayesian approach to system identification,'' in \emph{Trends and
  Progress in System Identification}, P.~Eykhoff, Ed.\hskip 1em plus 0.5em
  minus 0.4em\relax Oxford: Pergamon Press, 1981, pp. 239--304.

\bibitem{Kar:14Laz}
M.~K\'{a}rn\'{y}, K.~Macek, and T.~Guy, ``Lazy fully probabilistic design of
  decision strategies,'' in \emph{Advances in Neural Networks – ISNN 2014},
  Z.~Zeng, Y.~Li, and I.~King, Eds., International Symposium on Neural
  Networks.\hskip 1em plus 0.5em minus 0.4em\relax Springer, 2014, pp.
  140--149.

\bibitem{Fer74}
T.~Ferguson, ``Prior distributions on spaces of probability measures,''
  \emph{The Annals of Statistics}, vol.~2, no.~4, pp. 615--629, 1974.

\bibitem{Wat89}
C.~Watkins, ``Learning from delayed rewards,'' Ph.D. dissertation, King's
  College, Cambridge, May 1989.

\bibitem{Hum:16}
\BIBentryALTinterwordspacing
C.~Hummersone, ``Alternative box plot,'' 2016. [Online]. Available:
  \url{\url{https://www.github.com/IoSR-Surrey/MatlabToolbox}}
\BIBentrySTDinterwordspacing

\bibitem{Ull:94}
A.~Aho and J.~Ullman, \emph{Foundations of computer science}.\hskip 1em plus
  0.5em minus 0.4em\relax W.H. Freeman \& Co., 1994, ch. The Running Time of
  Programs.

\end{thebibliography}

\end{document}